\documentclass{article}

\usepackage[dblblindworkshop, final]{neurips_2025}
\usepackage[utf8]{inputenc} 
\usepackage[T1]{fontenc}    
\usepackage{hyperref}       
\usepackage{url}            
\usepackage{booktabs}       
\usepackage{amsfonts}       
\usepackage{nicefrac}       
\usepackage{microtype}      
\usepackage{xcolor}         
\usepackage{amsmath}
\usepackage{multirow}
\usepackage{graphicx}
\usepackage{subcaption}
\usepackage{bbm} 

\title{Spatio-Temporal Graphs Beyond Grids:\\Benchmark for Maritime Anomaly Detection}
\workshoptitle{AI for Science: The Reach and Limits of AI for Scientific Discovery}

\author{
  Jeehong Kim\thanks{Equal contribution (co-first authors).}\\
  Graduate School of Data Science\\
  Seoul National University\\
  \texttt{williamkim10@snu.ac.kr}
  \And
  Youngseok Hwang\footnotemark[1]\\
  Graduate School of Data Science\\
  Seoul National University\\
  \texttt{yshwang35@snu.ac.kr}
  \And
  Minchan Kim\\
  Graduate School of Data Science\\
  Seoul National University\\
  \texttt{mmm5373@snu.ac.kr}
  \And
  Sungho Bae\\
  Graduate School of Data Science\\
  Seoul National University\\
  \texttt{sunghobae@snu.ac.kr}
  \And
  Hyunwoo Park\\
  Graduate School of Data Science\\
  Seoul National University\\
  \texttt{hyunwoopark@snu.ac.kr}
}

\begin{document}

\maketitle

\begin{abstract}
Spatio-temporal graph neural networks (ST-GNNs) have achieved notable success in structured domains such as road traffic and public transportation, where spatial entities can be naturally represented as fixed nodes. In contrast, many real-world systems including maritime traffic lack such fixed anchors, making the construction of spatio-temporal graphs a fundamental challenge. Anomaly detection in these non-grid environments is particularly difficult due to the absence of canonical reference points, the sparsity and irregularity of trajectories, and the fact that anomalies may manifest at multiple granularities. In this work, we introduce a novel benchmark dataset for anomaly detection in the maritime domain, extending the Open Maritime Traffic Analysis Dataset (OMTAD) into a benchmark tailored for graph-based anomaly detection. Our dataset enables systematic evaluation across three different granularities: node-level, edge-level, and graph-level anomalies. We plan to employ two specialized LLM-based agents: \emph{Trajectory Synthesizer} and \emph{Anomaly Injector} to construct richer interaction contexts and generate semantically meaningful anomalies. We expect this benchmark to promote reproducibility and to foster methodological advances in anomaly detection for non-grid spatio-temporal systems. 
\end{abstract}
\section{Introduction}
\label{sec:intro}

Spatio-temporal graph neural networks (ST-GNNs) have been extensively studied in domains such as road traffic forecasting and public transportation systems~\cite{yu2017spatio,li2018diffusion,guo2019attention}. 
A common characteristic of these applications is that the underlying spatial entities like road intersections, bus stops, or subway stations can be naturally defined as fixed nodes. 
This inherent grid-like structure makes the construction of spatio-temporal graphs straightforward and facilitates the modeling of both spatial dependencies and temporal dynamics. 
Consequently, anomaly detection in such structured environments has received significant attention and demonstrated promising results~\cite{deng2021graph}.

However, there are many cases both in real-world and scientific domains where situations do not conform to these assumptions. In particular, there exist domains where fixed spatial anchors are absent or physically ambiguous. The maritime environment represents one of the most prominent examples: unlike road traffic systems, the open sea does not provide natural fixed nodes such as intersections or road segments. Although artificial proxies such as waypoints, port coordinates, or grid discretizations can be imposed, these methods are often ad hoc and fail to capture the continuous and dynamic nature of vessel trajectories. 
This fundamental challenge renders the construction of a meaningful spatio-temporal graph a non-trivial task. We expect that such \emph{non-grid spatio-temporal systems} will become increasingly common, not only in maritime monitoring but also in emerging domains such as drone swarms and aerial traffic management.

Performing anomaly detection in these settings is even more challenging. First, the lack of fixed spatial anchors complicates the definition of normal versus abnormal interactions among moving entities. Second, the inherent sparsity and irregularity of the trajectories make it difficult to design robust models. 
Third, anomalous patterns may manifest at multiple levels: individual entities (node-level anomalies), unusual pairwise interactions (edge-level anomalies), or entire subgroups behaving abnormally (graph-level anomalies). These challenges highlight the need for systematic benchmarks that enable rigorous evaluation and foster methodological innovations~\cite{ma2021comprehensive}.
There are several Marine datasets 

To address this gap, in this paper we introduce a novel benchmark dataset for anomaly detection in the maritime domain. 
Our dataset is designed to support anomaly detection tasks at three granularities: 
\emph{(i) node-level anomalies}, capturing abnormal single-entity behaviors, 
\emph{(ii) edge-level anomalies}, reflecting irregular inter-entity interactions, and 
\emph{(iii) graph-level anomalies}, identifying collective abnormal events. 
Inspired by recent advances in graph anomaly detection across node-, edge-, and graph-level settings~\cite{lin2024unigad}, we aim to provide a unified testbed that allows the community to explore and compare methods across multiple anomaly detection settings. 
To construct the dataset in a principled manner, we use two large language model(LLM)-based agents: \emph{Trajectory Synthesizer}, which augments inter-vessel contexts by enriching sparse neighborhoods, and an \emph{Anomaly Injector}, which introduces diverse anomalies guided by high-level prompts. 
We believe this contribution will not only facilitate research on maritime anomaly detection, but also establish a foundation for studying anomaly detection in broader non-grid spatio-temporal systems.

\section{Dataset}
We build our benchmark upon the \textbf{Open Maritime Traffic Analysis Dataset (OMTAD)}~\cite{masek2021omtad}, 
a publicly available and openly licensed collection of vessel trajectories derived from AIS signals. 
OMTAD covers the West Australian offshore region (105–116\textdegree E, 36–15\textdegree S) from 2018 to 2020, and provides \textbf{19,124} trajectories across four vessel categories: Cargo (14{,}384), Tanker (4{,}020), Fishing (466), and Passenger (254). 
Each AIS record includes vessel identifiers, geolocation, kinematic information such as course over ground (COG) and speed over ground (SOG), and UTC timestamps.

\subsection{Limitations and Our Extensions}
While OMTAD provides a well-organized and open collection of vessel tracks, it has two key limitations that prevent direct use for graph-based anomaly detection. First, although some trajectories are physically close enough to form meaningful spatio-temporal graphs, many occur in isolation without nearby neighbors, making graph construction difficult. Second, it only contains \emph{normal} trajectories and thus provides no anomaly labels. These issues hinder systematic benchmarking of graph-based anomaly detection.

To address these limitations, we extend OMTAD in two complementary ways. 
\begin{itemize}
    \item \textbf{Trajectory synthesis in sparse regions.} For vessels without nearby neighbors, we generate synthetic but physically plausible companion trajectories. These synthetic neighbors are created by perturbing SOG, COG, and geolocation values within bounded ranges, ensuring that even isolated vessels can be embedded into meaningful spatio-temporal graphs. 
    \item \textbf{Anomaly injection.} Since no anomalies are provided, we introduce anomalies through a controlled injection process. Instead of rigid rules, we rely on prompt-driven generation to produce diverse anomalies across node, edge, and graph levels, aligning them with semantically meaningful maritime scenarios. 
\end{itemize}

This extension is not only practical but also justified: preliminary experiments in Appendix~\ref{sec:Preliminary Experiment} show that even under relatively naive anomaly settings, graph-based models consistently outperform purely temporal baselines. This validates that repurposing OMTAD into a graph-based anomaly detection benchmark captures meaningful structural signals and provides a solid foundation for further extensions. Through these steps, OMTAD is repurposed into a unified benchmark dataset that supports systematic evaluation of anomaly detection in non-grid spatio-temporal graphs. Detailed configurations of the trajectory synthesis and anomaly injection agents are provided in the Appendix~\ref{sec:Dataset Construction Method}.

\bibliographystyle{abbrv}
\bibliography{references}

\appendix

\section{Motivation}
To verify the feasibility of our proposed benchmark, we conducted a preliminary experiment by injecting synthetic anomalies into the OMTAD dataset. 
This was necessary because maritime data lacks ground-truth anomaly labels and defining anomalies is highly context-dependent, with no clear consensus even within the maritime community. 
By introducing controlled perturbations, we created a testbed to examine whether graph-level anomaly detection tasks can be meaningfully supported in this setting.

\subsection{Anomaly Injection}
We synthesized anomalies by perturbing vessel trajectories at the node level. 
For each trajectory of length $w$, a contiguous anomaly block of size $m = r_{\text{node}} w$ was chosen, where $r_{\text{node}} \in \{r_1, r_2, r_3\}$ is the node anomaly ratio. 
The block was placed by sampling a start index $s \sim \mathcal{U}(0, w-m)$, which defined a binary anomaly mask $z_t$ indicating anomalous segments. 
Nodes within the anomaly block were perturbed in their Speed Over Ground (SOG) and Course Over Ground (COG) features~\cite{liu2024ais}. 
Formally, we modeled the rates of change of SOG ($a$) and COG ($\omega$) as normally distributed, 
$a \sim \mathcal{N}(\mu_a, \sigma_a^2)$ and $\omega \sim \mathcal{N}(\mu_\omega, \sigma_\omega^2)$, 
where $a_i = (\mathrm{SOG}_i - \mathrm{SOG}_{i-1}) / \Delta t$ and 
$\omega_i = (\mathrm{COG}_i - \mathrm{COG}_{i-1}) / \Delta t$. 
To create significant deviations, we replaced them with 
$a_i^* = \mu_a + k \cdot \sigma_a$ and $\omega_i^* = \mu_\omega + k \cdot \sigma_\omega$ with $k > 3$,
which ensures perturbed values lie outside the 99.7\% confidence interval of normal behavior. 
The updated SOG and COG values were then iteratively applied over the anomaly block. 
A trajectory was labeled anomalous ($y_{\mathrm{traj}}=1$) if at least one node was perturbed, and normal ($y_{\mathrm{traj}}=0$) otherwise. As mentioned earlier, defining anomalies in the maritime context is inherently difficult, and even within this domain there is no established consensus. Therefore, we restrict our anomaly definition to \emph{kinematic movement anomalies} based on SOG and COG deviations. 

\paragraph{Node- and graph-level anomaly ratios.}
In our design, anomaly prevalence is controlled at two complementary levels. 
First, the \emph{node anomaly ratio} $r_{\text{node}} \in (0,1]$ specifies the fraction of anomalous nodes within a trajectory. 
Given a trajectory of length $w$, the anomalous span is set to $m = r_{\text{node}} \, w$ nodes, realized as a consecutive block of length $m$. 
This choice reflects the temporal persistence of real-world incidents, since anomalies are more likely to appear as sustained abnormal behaviors (e.g., equipment malfunction, evasive maneuvers, adverse weather conditions, or loitering) rather than isolated spikes.

Second, the \emph{trajectory anomaly ratio} $r_{\text{traj}} \in (0,1]$ denotes the fraction of trajectories labeled anomalous at the graph level, formally defined as

\[
r_{\text{traj}} \;=\; \frac{1}{N} \sum_{i=1}^{N} \mathbbm{1}\!\left( y_{\mathrm{traj}}^{(i)} = 1 \right),
\]

where $N$ is the total number of trajectories. 

Thus, $r_{\text{node}}$ controls the intra-trajectory anomaly density, while $r_{\text{traj}}$ governs the dataset-level class balance for graph-level detection.



\subsection{Preliminary Experiment}
\label{sec:Preliminary Experiment}
\begin{figure}[!h]
    \centering
    \begin{subfigure}[t]{0.48\linewidth}
        \centering
        \includegraphics[width=\linewidth]{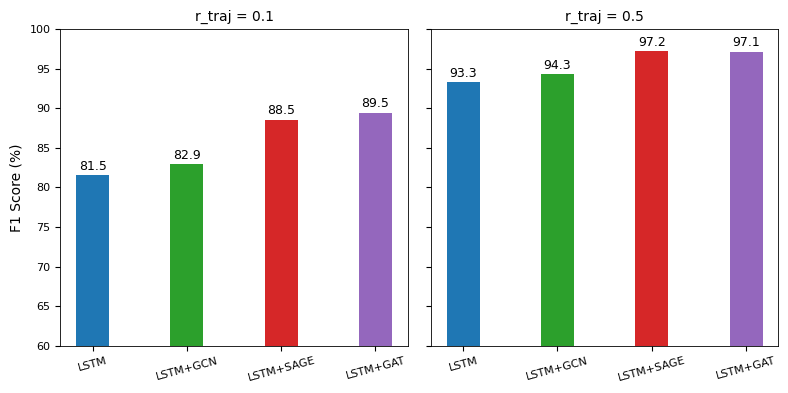}
        \caption{}
        \label{fig:prelim-lstm}
    \end{subfigure}
    \hfill
    \begin{subfigure}[t]{0.48\linewidth}
        \centering
        \includegraphics[width=\linewidth]{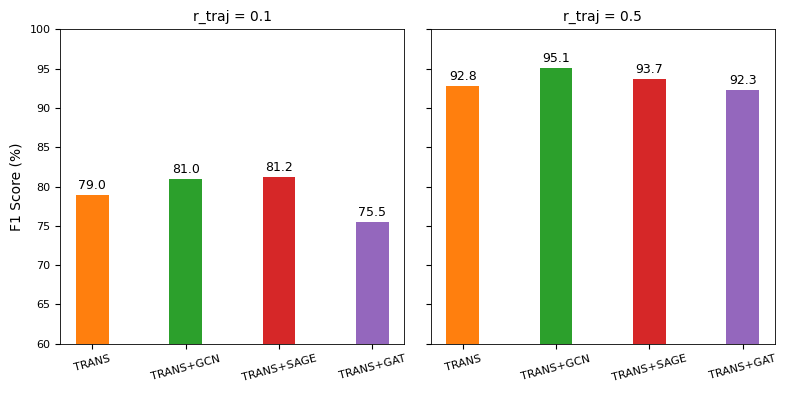}
        \caption{}
        \label{fig:prelim-trans}
    \end{subfigure}
    \caption{Preliminary results of time-series models and their GNN-integrated variants under different $r_{\text{traj}}$ settings. (a) LSTM-based models. (b) Transformer-based models. ``TRANS'' denotes Transformer.}
    \label{fig:prelim-combined}
\end{figure}

We conducted a preliminary study under the setting of \emph{graph-level anomaly detection}, where each vessel trajectory is classified as either normal or anomalous. 
In this setup, we varied the trajectory anomaly ratio $r_{\text{traj}} \in \{0.1, 0.5\}$ while fixing the node anomaly ratio at $r_{\text{node}}=0.5$. 
We compared standard time-series models (LSTM, Transformer) with their \emph{Time-series + GNN} counterparts, which incorporate GNN modules, as summarized in Fig.~\ref{fig:prelim-lstm} and Fig.~\ref{fig:prelim-trans}.

To construct graph inputs for the GNNs, we applied the OPTICS clustering algorithm to spatial snapshots at each timestamp $t$, grouping nearby vessels into dynamic clusters without imposing predefined constraints on the number or shape of clusters. 
From each cluster, a fixed number $k$ of vessel trajectories was sampled to ensure that the adjacency matrix remained consistent across all graphs. 
For each sampled set, we then built a directed temporal graph over a $w$-hour observation window, where each node corresponds to a time-stamped vessel state, resulting in exactly $k \times w$ nodes per graph. 

Although this preliminary experiment primarily focuses on \emph{graph-level} anomaly detection with varying $r_{\text{traj}}$ and fixed $r_{\text{node}}$, the same framework can naturally be extended to support \emph{node-level} anomaly detection.

\subsection{Findings}
The results show that GNN-integrated models consistently outperform purely temporal baselines across all anomaly ratios. 
This demonstrates that graph-based modeling provides a more natural fit for capturing maritime dynamics, where vessel states and inter-vessel interactions must be considered jointly. 
At the same time, it is important to emphasize that our injection strategy perturbed only the simplest navigational features. Real-world maritime anomalies are far more diverse, including illegal rendezvous between vessels, loitering behaviors, spoofed AIS signals, or sudden deviations due to environmental conditions. 
Thus, while this experiment confirmed the viability of our framework, it represents only a simplified case. 
Our ultimate goal is to generalize this process by employing LLM-based agents to automatically generate and annotate richer, semantically meaningful anomalies, thereby creating a more realistic and versatile benchmark environment.

\section{Related Works}
\paragraph{Spatio-temporal GNNs in structured domains.}
Spatio-temporal graph neural networks (ST-GNNs) have achieved notable success in domains where the underlying graph structure is fixed and well defined, such as road traffic~\cite{yu2017spatio,li2018diffusion,wu2019graph}, public transportation~\cite{guo2019attention}, and mobility systems~\cite{lim2020stp,deng2021graph}. 
In these settings, nodes typically correspond to pre-defined spatial anchors (e.g., intersections, stations, or sensors), which makes the construction of spatio-temporal graphs straightforward and effective. 
However, the assumptions of stable topologies and fixed node identities do not generalize to non-grid environments such as the open sea, where spatial anchors are absent and trajectories are highly irregular. 


\paragraph{Maritime Anomaly Detection and Datasets.}
Maritime anomaly detection has emerged as a challenging task in the maritime domain due to the dynamic and unstructured nature of vessel movements. A comprehensive survey~\cite{riveiro2018maritime} highlights the difficulty of defining anomalous behaviors and reviews a wide range of approaches. Classical machine learning techniques have long been applied to AIS data, including supervised and unsupervised methods for identifying irregular navigation patterns~\cite{obradovic2014machine,singh2020machine}. More recently, deep learning approaches have demonstrated stronger capacity for modeling complex temporal dependencies, such as CNN- and RNN-based models for abnormal behavior detection~\cite{zang2021abnormal} and probabilistic neural representations like GeoTrackNet~\cite{nguyen2021geotracknet}. Transformer-based methods have also been introduced, with TrAISformer~\cite{nguyen2024transformer} achieving state-of-the-art results in AIS-based trajectory prediction. In parallel, graph-based methods have gained momentum for their ability to explicitly capture vessel-to-vessel interactions, including graph attention networks for anomaly detection~\cite{zhang2023vessel} and spatio-temporal graph convolutional networks for trajectory prediction in crowded sea areas~\cite{wang2024vessel}. 

A key bottleneck in advancing this line of research lies in the lack of standardized open datasets. While several AIS-based datasets exist~\cite{mao2017automatic}, they are often incomplete, commercial, or unavailable for public use. The Open Maritime Traffic Analysis Dataset (OMTAD)~\cite{masek2021omtad} represents an important step toward openness by providing cleaned and processed AIS tracks for multiple vessel types. Nevertheless, OMTAD has not been designed as an anomaly detection benchmark, and in particular, it lacks systematic definitions and annotations for multi-level anomalies. Our work addresses this gap by extending OMTAD into a benchmark dataset tailored for graph-based anomaly detection across node, edge, and graph levels.


\paragraph{LLM-based Anomaly Injection and Benchmark Augmentation.}
Recent studies have begun to explore the potential of LLMs in supporting anomaly detection tasks. For instance, AD-LLM~\cite{yang2024ad} presents the first comprehensive benchmark that systematically examines how LLMs can be leveraged for anomaly detection across multiple dimensions, including zero-shot detection, data augmentation, and model selection. This line of work demonstrates the broad applicability of LLMs in enhancing anomaly detection pipelines. However, these efforts primarily remain at an abstract level and provide limited insights into fine-grained dataset augmentation grounded in real-world domain data. 

In parallel, BotSim~\cite{qiao2025botsim} introduces an LLM-powered end-to-end simulation toolkit for malicious social botnet generation, enabling downstream evaluation of bot detection methods. 
This framework illustrates how LLM agents can be utilized to construct diverse and semantically meaningful anomaly scenarios in a simulation setting. 
Nevertheless, the maritime domain remains underexplored: despite the availability of AIS-based datasets such as OMTAD, there has been little research on using LLMs to perform precise, domain knowledge–driven anomaly injection.

To bridge this gap, we extend OMTAD into a benchmark dataset specifically designed for maritime anomaly detection. To the best of our knowledge, this is the first attempt to systematically augment a real-world maritime dataset with LLM-based anomaly injection, providing a platform for training and evaluating anomaly detection methods in non-grid spatio-temporal systems.

\section{Dataset Construction Method}
\label{sec:Dataset Construction Method}
\paragraph{Overview}
We adopt a two–agent architecture specialized for dataset construction: 
(1) \emph{Trajectory Synthesizer}, which enriches inter-vessel connectivity through augmentation of local contexts, and 
(2) \emph{Anomaly Injector}, which introduces anomalies guided by high-level text prompts. 
Both agents operate under a common \emph{Coordinator} that manages data flow, prepares structured perception inputs, enforces constraints, and validates outputs. 
This design separates augmentation (ensuring sufficient structural density) from anomaly generation (ensuring semantic variety), providing a flexible and reproducible pipeline for benchmark creation.

\subsection{Coordinator Workflow}
For each focal vessel $v$ over a given window $[t_0,t_1]$, the Coordinator executes a simple loop: 
(i) construct a standardized \emph{perception bundle} from AIS and environmental metadata, 
(ii) dispatch it to the Trajectory Synthesizer to obtain an augmented multi-vessel graph $\mathcal{G}$, 
(iii) pass the synthesized graph and perception context to the Anomaly Injector to apply prompt-driven modifications and produce labels, and 
(iv) collect provenance, validation logs, and final artifacts for dataset assembly. 
In this way, augmentation and anomaly injection are decoupled but remain interoperable under a single orchestrator.

\subsection{Shared Environment Perception Schema}
Both agents consume a common schema that represents vessel states and their context in a slot-filled format. The specific categories and fields are mentioned in Table~\ref{tab:perception-schema}.
\begin{table}[t]
\centering
\caption{Perception schema consumed by both agents.}
\label{tab:perception-schema}
\begin{tabular}{ll}
\toprule
\textbf{Category} & \textbf{Fields} \\
\midrule
AIS & MMSI, $t$, latitude, longitude, SOG, COG\\
Derived & $\Delta$SOG/$\Delta t$, $\Delta$COG/$\Delta t$\\
Env & wind/wave/current bins, visibility proxy \\
Provenance & source trajectory IDs \\
\bottomrule
\end{tabular}
\end{table}

This schema ensures that both augmentation and injection modules operate on consistent, validated inputs. 
All fields follow fixed units and identifiers, and missing values are explicitly marked to maintain determinism.

\subsubsection{Agent~1: Trajectory Synthesizer (Augmentation)}
\textbf{Goal.} Increase the density and diversity of meaningful inter-vessel interactions so that GNN-based methods can better exploit spatial context while preserving physical plausibility.

\textbf{Main Idea.} 
The Trajectory Synthesizer enriches local graph structures by adding trajectories around each vessel to ensure sufficient connectivity and realistic interaction density.

\textbf{Components.}
\begin{itemize}
  \item \textbf{Neighbor-based augmentation:} If physically close vessels are present, their trajectories are directly included to form proximity-based edges and enrich inter-vessel connectivity. 
  \item \textbf{Synthetic augmentation:} In sparse regions where nearby vessels are absent, the agent generates additional “virtual neighbors” by sampling trajectories similar to the focal vessel. 
  Their SOG, COG, latitude, and longitude values are perturbed within realistic variation ranges to preserve plausibility while increasing graph density. 
\end{itemize}

\textbf{Outputs.} 
An augmented spatio-temporal graph that combines original vessel tracks with either actual or synthesized neighbors, including provenance information indicating which trajectories were real and which were generated.

\subsubsection{Agent~2: Anomaly Injector (Prompt-Driven)}
\textbf{Goal.} Introduce diverse and semantically meaningful anomalies into trajectories in order to support node-, edge-, and graph-level anomaly detection tasks.

\textbf{Main Idea.} 
The Anomaly Injector operates from high-level \emph{text prompts} rather than fixed perturbation rules, allowing flexible and context-aware anomaly creation.

\textbf{Components.}
\begin{itemize}
  \item \textbf{Prompt Interpretation:} Parsing natural language descriptions of anomalies 
        (e.g., unusual speed changes, risky encounters, or group loitering) into structured intent. 
  \item \textbf{Scenario Realization:} Mapping the interpreted intent into corresponding edits of the spatio-temporal graph, 
        such as modifying single-node kinematics, vessel-to-vessel interactions, or group-level patterns. 
  \item \textbf{Label Generation:} Attaching anomaly labels (node, edge, or graph level) along with 
        rationale text that traces back to the original prompt. 
\end{itemize}

\textbf{Outputs.} 
A set of modified trajectories and anomaly labels, where each label is tied to a prompt, anomaly type, and severity level, accompanied by rationale text for interpretability.

\section{Future Works}
While our current work lays the foundation for a benchmark on non-grid spatio-temporal anomaly detection, several important directions remain for future development. First, we plan to consolidate the proposed pipeline into a reproducible framework that can automatically synthesize augmented trajectories and inject anomalies through prompt-driven agents. The next step is to curate a finalized version of the dataset. We will release the dataset under an open license to encourage broad adoption and reproducibility, accompanied by scripts that enable researchers to regenerate augmented or injected variants deterministically. 

Second, to establish a reference point for the community, we will benchmark a variety of baseline methods on the dataset. This includes purely temporal sequence models such as LSTM and Transformer, hybrid spatio-temporal GNN models, and recent graph anomaly detection architectures designed for node-, edge-, and graph-level tasks. Comprehensive evaluation across different anomaly ratios and scenarios will provide insights into the strengths and limitations of each model class. 

Finally, we aim to extend the anomaly definitions beyond the initial kinematic-focused injections. In particular, we plan to incorporate more semantically complex anomalies, such as illegal encounters, coordinated group behaviors, or procedural violations near ports and restricted areas. Leveraging LLM-based agents in conjunction with domain rules will allow us to gradually expand the scope of the benchmark, bridging the gap between controlled synthetic anomalies and realistic, context-dependent maritime events. In parallel, we recognize that the task-specific labeling strategy itself requires careful refinement. Defining consistent and interpretable labels across node-, edge-, and graph-level tasks is non-trivial, and we plan to investigate principled ways of assigning task-aware labels that capture both local anomalies and their broader contextual implications.

\end{document}